\documentclass[conference]{IEEEtran}
\IEEEoverridecommandlockouts

\usepackage{amsmath,amssymb,amsthm}
\usepackage{graphicx}
\usepackage{booktabs}
\usepackage{multirow}
\usepackage{float}
\usepackage{hyperref}
\usepackage{url}
\usepackage{caption}
\usepackage{subcaption}
\usepackage{mathtools}
\usepackage{algorithm}
\usepackage{algpseudocode}
\usepackage{enumitem}

\begin{document}

\title{SpectralKrum: A Spectral-Geometric Defense Against Byzantine Attacks in Federated Learning%
\thanks{Code is available at: \url{https://github.com/EdddTri/Spectral_Krum}}
}

\author{
    \IEEEauthorblockN{Aditya Tripathi}
    \IEEEauthorblockA{DA-IICT \\ Gandhinagar, India \\ 202318046@dau.ac.in}
    \and
    \IEEEauthorblockN{Karan Sharma}
    \IEEEauthorblockA{DA-IICT \\ Gandhinagar, India \\ 202318018@dau.ac.in}
    \and
    \IEEEauthorblockN{Rahul Mishra}
    \IEEEauthorblockA{IIT Patna \\ Patna, India \\ rahul\_mishra@iitp.ac.in}
    \and
    \IEEEauthorblockN{Tapas Kumar Maiti}
    \IEEEauthorblockA{DA-IICT \\ Gandhinagar, India \\ tapas\_kumar@dau.ac.in}
}

\maketitle

\begin{abstract}
Federated Learning (FL) distributes model training across clients who retain their data locally, but this architecture exposes a fundamental vulnerability: \emph{Byzantine} clients can inject arbitrarily corrupted updates that degrade or subvert the global model. While robust aggregation methods (including Krum, Bulyan, and coordinate-wise defenses) offer theoretical guarantees under idealized assumptions, their effectiveness erodes substantially when client data distributions are heterogeneous (non-IID) and adversaries can observe or approximate the defense mechanism.

This paper introduces \textbf{SpectralKrum}, a defense that fuses spectral subspace estimation with geometric neighbor-based selection. The core insight is that benign optimization trajectories, despite per-client heterogeneity, concentrate near a low-dimensional manifold that can be estimated from historical aggregates. SpectralKrum projects incoming updates into this learned subspace, applies Krum selection in the compressed coordinates, and filters candidates whose orthogonal residual energy exceeds a data-driven threshold. The method requires no auxiliary data, operates entirely on model updates, and preserves FL's privacy properties.

We evaluate SpectralKrum against eight robust baselines across seven attack scenarios on CIFAR-10 with Dirichlet-distributed non-IID partitions ($\alpha=0.1$). Our experiments, spanning over 56{,}000 training rounds, reveal a nuanced picture: SpectralKrum demonstrates competitive robustness against directional and subspace-aware attacks (adaptive-steer, buffer-drift) but provides limited advantage over simpler statistical methods under label-flip and min-max perturbations where malicious updates remain spectrally indistinguishable from benign ones. Rather than claiming universal superiority, we characterize precisely when spectral geometry aids Byzantine defense and document failure modes that inform future research directions.
\end{abstract}

\section{Introduction}

Federated Learning (FL) partitions model training across a network of clients, each holding private data that never leaves its device~\cite{mcmahan2017}. The server aggregates locally computed updates (typically gradients or parameter deltas) to refine a shared global model. This paradigm powers applications from mobile keyboard prediction~\cite{hard2018} to cross-institutional medical imaging~\cite{rieke2020}, but it introduces a critical security challenge: the server cannot verify the provenance or correctness of any individual update.

A \emph{Byzantine} client, whether compromised by an adversary or simply malfunctioning, may submit arbitrary vectors as its contribution. With even a modest fraction of Byzantine participants, naive averaging fails catastrophically: the global model can diverge, suffer degraded accuracy, or acquire targeted backdoor behaviors that activate only on adversary-chosen inputs.

\subsection{The Geometric Problem Under Non-IID Data}

Most robust aggregation guarantees assume that honest clients' updates cluster tightly around a shared mean. This assumption rarely holds in practice~\cite{li2020}. Real FL deployments exhibit pronounced data heterogeneity: a user's smartphone reflects their particular app usage patterns; a hospital's imaging data skews toward its regional patient demographics. Formally, when client data follows a Dirichlet distribution with concentration $\alpha \ll 1$, individual clients may observe only a subset of classes, producing gradients that point in legitimately different directions.

This heterogeneity fundamentally complicates Byzantine defense:

\begin{itemize}[leftmargin=*]
    \item Benign gradients spread across parameter space, potentially spanning multiple distinct modes.
    \item Distance-based methods like Krum, designed to select the update nearest its neighbors, may oscillate between benign subgroups or mistake a centrally-positioned attack for the honest consensus.
    \item Coordinate-wise defenses (trimmed mean, median) ignore correlations between parameters, allowing adversaries to craft perturbations that are small in each coordinate but large in aggregate.
\end{itemize}

The geometric intuition behind classical defenses thus breaks down precisely when it is most needed.

\subsection{Spectral Structure as a Defense Signal}

Despite this heterogeneity, FL optimization exhibits a remarkable regularity: aggregated updates from honest clients, viewed across multiple rounds, trace a trajectory that lies near a low-dimensional manifold~\cite{li2018measuring}. The dominant principal components of this trajectory capture directions of shared descent, while noise and client-specific variation concentrate in the residual subspace.

This observation suggests a complementary defense signal. An update that deviates substantially from the historical subspace, measured by its \emph{orthogonal energy}, is geometrically anomalous, regardless of how close it may appear to any single neighbor. Spectral methods like DnC-PMF exploit this insight, but they remain vulnerable to adversaries who deliberately project their malicious updates into the benign subspace, matching the spectral signature while steering the model in a harmful direction.

\subsection{Contributions and Scope}

We propose \textbf{SpectralKrum}, a defense that integrates spectral subspace estimation with Krum's neighbor-based selection. The method:

\begin{enumerate}[leftmargin=*]
    \item Maintains a rolling buffer of past aggregated updates and constructs a trimmed PCA subspace capturing benign optimization dynamics.
    \item Projects incoming client updates into this subspace and applies Krum selection in the compressed, low-dimensional coordinates where benign clustering is tighter.
    \item Filters the Krum-selected candidates using an orthogonal energy threshold calibrated from the historical distribution of residuals.
\end{enumerate}

Crucially, SpectralKrum operates without access to client data, labels, or any trusted root dataset, preserving FL's core privacy guarantees.

Our evaluation on CIFAR-10 with highly non-IID Dirichlet partitions ($\alpha=0.1$) reveals both the promise and the limits of this approach. SpectralKrum matches or exceeds baseline robustness against directional attacks (sign-flip) and subspace-aware adversaries (adaptive-steer, buffer-drift) that introduce detectable spectral anomalies. However, it provides no consistent advantage over simpler statistical aggregators when malicious updates remain spectrally indistinguishable from benign ones, as occurs with label-flip poisoning and carefully calibrated min-max perturbations.

Rather than claiming a universal solution, we contribute:
\begin{itemize}[leftmargin=*]
    \item A principled algorithm combining spectral and geometric robustness.
    \item A rigorous empirical study across seven attack families and eight robust baselines.
    \item A characterization of when spectral defenses succeed, when they fail, and what gaps remain for future work.
\end{itemize}

\section{Related Work}

The challenge of robust aggregation in distributed learning intersects several research threads: Byzantine-tolerant distributed optimization, robust statistics, adversarial machine learning, and dimensionality reduction. We review these areas with attention to how each informs, and where each falls short of, the non-IID FL setting.

\subsection{Byzantine-Tolerant Distributed Optimization}

The theoretical foundations of Byzantine robustness trace to distributed consensus problems, where the goal is to compute a function of honest participants' inputs despite arbitrary failures. Lamport's original formulation~\cite{lamport1982} established that tolerating $f$ Byzantine nodes requires at least $3f+1$ total participants under certain communication models. In the ML setting, the analogous question is whether robust gradient descent can converge despite corrupted updates.

\subsubsection{Coordinate-wise Aggregation}
The simplest robust estimators operate independently on each parameter coordinate. \emph{Coordinate Median}~\cite{tukey1960} replaces averaging with the median; \emph{Trimmed Mean}~\cite{yin2018} discards the largest and smallest fraction of values before averaging. Both inherit the breakdown point~\cite{donoho1983} of their scalar counterparts: they tolerate up to $\lfloor n/2 \rfloor - 1$ adversarial values per coordinate. Recent work on robust mean estimation~\cite{pillutla2022} further refines these guarantees.

The limitation is fundamental. By treating coordinates independently, these methods ignore correlations in the gradient structure. An adversary can construct an attack vector that is small in every coordinate yet large in norm, escaping coordinate-wise detection while significantly perturbing the aggregate. Under non-IID conditions, this vulnerability is amplified: legitimate coordinate-wise variation already approaches the detection threshold, leaving little margin for filtering attacks.

\subsubsection{Geometric Median}
The geometric median $\hat{\mu} = \arg\min_{v} \sum_i \|v - d_i\|_2$ provides a dimension-independent robustness guarantee~\cite{chen2017} and is a natural choice when coordinate independence is untenable. Weiszfeld's algorithm~\cite{weiszfeld1937} gives an efficient iterative solution in most cases.

In practice, however, the geometric median suffers from two drawbacks in FL. First, when benign updates form multiple clusters (as under non-IID data), the geometric median converges to a ``compromise'' point that may represent no client's actual descent direction, slowing convergence. Second, iterative computation is expensive for the high-dimensional parameter vectors typical of modern neural networks.

\subsubsection{Krum and Multi-Krum}
Blanchard et al.~\cite{blanchard2017} introduced Krum as a selection-based alternative: rather than averaging, choose the single update whose sum of distances to its $n - f - 2$ nearest neighbors is smallest. The intuition is that honest updates cluster, while Byzantine updates are isolated or collusive outliers. Multi-Krum generalizes this by selecting the top-$k$ candidates and averaging them.

Krum's guarantees assume tight benign clustering, precisely the condition violated under non-IID data. When benign updates spread across multiple modes, Krum may oscillate between modes or select an adversary positioned at the geometric center of the dispersed benign set. Our experiments confirm this instability: FullKrum achieves only 26--28\% mean accuracy across attacks (Table~\ref{tab:auc_summary}), worse than simple coordinate medians.

\subsubsection{Bulyan}
Mhamdi et al.~\cite{mhamdi2018} proposed Bulyan as a two-phase defense: iteratively apply Krum to select $n - 2f$ candidates, then apply coordinate-wise trimmed mean to the selection. This composition inherits some robustness from both components, but also their limitations: Krum's sensitivity to non-IID dispersion and trimmed mean's coordinate independence. Bulyan additionally requires $n \geq 4f + 3$ participants, a constraint that may not hold in small cohorts.

\subsection{Spectral and Subspace Methods}

A distinct line of work observes that benign gradient trajectories occupy a low-dimensional subspace, motivating defenses that detect anomalies in spectral coordinates.

\subsubsection{DnC-PMF and DnC-Cluster}
Shejwalkar and Houmansadr~\cite{shejwalkar2021} introduced Divide-and-Conquer defenses that project updates into a PCA subspace, cluster them (DnC-Cluster) or filter by projection magnitude (DnC-PMF), and aggregate only the selected subset. These methods can reject attacks that introduce variance orthogonal to the benign subspace.

The vulnerability is equally clear: an adversary with knowledge of the subspace can project its malicious update \emph{into} that subspace, evading detection. Our adaptive-steer attack explicitly implements this strategy, demonstrating that spectral consistency alone is insufficient.

\subsubsection{Historical Subspace Estimation}
Rather than estimating the subspace from the current round's updates (which may include adversaries), SpectralKrum uses a rolling buffer of \emph{past aggregates}, each presumably less contaminated because it is itself the output of robust aggregation. This temporal smoothing yields a more stable subspace estimate, though it introduces latency: the subspace adapts slowly to legitimate distributional shifts.

\subsection{Model Poisoning and Backdoor Attacks}

Byzantine attacks in FL fall into two broad categories. \emph{Untargeted} (or model-poisoning) attacks aim to degrade overall accuracy~\cite{sun2020}; examples include sign-flip, where the adversary submits $-\gamma \cdot g$ for honest gradient $g$, and min-max~\cite{fang2020}, where the adversary computes the worst-case perturbation within the benign variance envelope.

\emph{Targeted} (or backdoor) attacks~\cite{bagdasaryan2020} seek to implant behavior triggered by specific inputs while preserving accuracy on clean data. A semantic backdoor embeds a visual trigger (e.g., a pixel patch) in training images, relabeling them to a chosen target class. Because the resulting gradient points in a legitimately useful direction for the trigger distribution, it may exhibit low orthogonal energy and evade spectral filters.

\subsection{Defenses Requiring Trusted Data}

Several recent defenses assume access to a small, clean ``root'' dataset held by the server. FLTrust~\cite{cao2021} computes a trusted gradient on this root data and re-weights client updates by their cosine similarity to the trusted direction. While effective, this assumption is often impractical in cross-device FL~\cite{kairouz2021}, where the server has no labeled data and acquiring such data undermines the privacy motivation of federation.

SpectralKrum avoids this assumption entirely: it operates solely on the model updates, requiring no auxiliary data or labels.

\subsection{Positioning SpectralKrum}

Existing defenses each address part of the Byzantine problem:
\begin{itemize}[leftmargin=*]
    \item Coordinate-wise methods handle scalar outliers but ignore directionality.
    \item Geometric methods (Krum, median) exploit clustering but fail under non-IID dispersion.
    \item Spectral methods detect orthogonal anomalies but succumb to subspace-aware imitation.
\end{itemize}

SpectralKrum combines spectral projection (to tighten benign clustering and detect orthogonal anomalies) with Krum selection (to exploit neighbor structure in the compressed space). Our experiments test whether this combination yields benefits beyond its components, and under what conditions it does not.

\section{Method: SpectralKrum}

SpectralKrum addresses a fundamental tension in Byzantine-robust FL: geometric defenses like Krum assume tight clustering of honest updates, while spectral defenses like DnC assume anomalies introduce orthogonal variance. Neither assumption holds reliably under non-IID data with sophisticated adversaries. Our approach combines both signals: applying Krum in a learned spectral subspace where clustering is tighter, while filtering by orthogonal energy to catch updates that evade geometric selection.

\subsection{Problem Setting}

Each FL round, the server receives updates $\{d_i\}_{i=1}^n \subset \mathbb{R}^d$ from $n$ clients, of which up to $f$ may be Byzantine. The goal is to compute an aggregate $a$ that approximates the honest mean. SpectralKrum achieves this through four stages: (1) estimate a benign subspace from historical aggregates, (2) project current updates into this subspace, (3) run Krum selection in the compressed coordinates, and (4) filter by orthogonal energy before averaging.

\subsection{Historical Subspace Estimation}

A key design choice in SpectralKrum is estimating the benign subspace from \emph{past aggregates} rather than current-round updates. Let $\mathcal{B} = \{g^{(t-B+1)}, \ldots, g^{(t)}\}$ denote the rolling buffer of the $B$ most recent aggregated updates. Since each $g^{(\cdot)}$ is itself the output of robust aggregation, the buffer is less contaminated than raw client submissions.

Stacking the buffer as $X \in \mathbb{R}^{B \times d}$, we center by subtracting the coordinate-wise mean (or median for additional robustness) and trim rows by $\ell_2$ norm to reduce the influence of any past aggregates that may have been adversarially biased. Specifically, we discard the $\alpha$-fraction of rows with the largest and smallest norms, retaining the central $(1-2\alpha)$ fraction.

\subsection{PCA Subspace Construction}

On the trimmed, centered buffer $X_k$, we compute a rank-$r$ PCA~\cite{johnstone2001}:
\[
U = \text{PCA}(X_k, r) \in \mathbb{R}^{d \times r}, \quad U^\top U = I_r.
\]
The columns of $U$ span the estimated benign subspace, and $P = UU^\top$ is the corresponding projection operator. The trimming step follows robust estimation principles~\cite{andrews2003}. If the buffer is too small ($|X_k| < 2$) or $r$ exceeds the available rank, we fall back to coordinate-wise median aggregation for that round.

\subsection{Orthogonal Energy and Threshold Calibration}

For any update $d$, define its orthogonal energy as the norm of its component outside the benign subspace:
\[
\rho(d) = \|d - UU^\top d\|_2.
\]
Intuitively, $\rho(d)$ measures how much of $d$ lies in directions not captured by benign optimization history. We calibrate a threshold $\tau$ as the $q$-th quantile (default $q=0.98$) of orthogonal energies computed over the buffer:
\[
\tau = \mathrm{quantile}_q\bigl(\{\rho(g^{(j)})\}_{j=1}^B\bigr).
\]

\subsection{Projection and Selection}

Given incoming updates $\{d_i\}$, we project each to spectral coordinates $z_i = U^\top d_i \in \mathbb{R}^r$. In this compressed space, we compute pairwise distances and apply Krum: select the $k = n - f - 2$ updates whose sum of distances to their $n-f-2$ nearest neighbors is smallest. Denote this set as $S$.

\subsection{Krum in Spectral Coordinates}

Instead of running Krum on full-dimensional updates, SpectralKrum applies Krum selection to the low-dimensional projections $\{z_i\}$:

\[
S = \text{KrumSelect}(\{z_i\}, f).
\]

In low dimensions, benign updates cluster more tightly, adversarial directionality is suppressed, and distance computations avoid the pathologies of high-dimensional concentration~\cite{beyer1999}. Reliability depends on the accuracy of the learned subspace and the adversary's sophistication.

\subsection{Guard Filtering}

We filter $S$ by orthogonal energy: $G = \{i \in S : \rho(d_i) \le \tau\}$.

If $G$ becomes too small (e.g., attackers positioned themselves in subspace), we 
retain the minimal-residual candidates:

\[
G = \arg \min_{\substack{T \subseteq S \\ |T|=k}} \sum_{i\in T} \rho(d_i),
\]
where $k = \text{guard\_min\_kept}$.

This fallback ensures aggregation continues even when Krum selects suspicious candidates.
Adaptive subspace-aware adversaries may still construct updates whose orthogonal energy lies within the benign threshold, limiting the filter's effectiveness.

\subsection{Aggregation}

Finally, compute:

\[
a = \frac{1}{|G|} \sum_{i\in G} d_i.
\]

This aggregated update is appended to the buffer (evicting the oldest entry), 
closing the loop.

\subsection{Full Algorithm Pseudocode}

\begin{algorithm}[h]
\caption{SpectralKrum Algorithm}
\begin{algorithmic}[1]

\State \textbf{Input:} updates $d_1,\ldots,d_n$, buffer $X$, PCA dimension $r$, 
trim fraction $\alpha$, quantile $q$, Krum Byzantine parameter $f$
\State \textbf{Output:} aggregated update $a$

\vspace{0.3em}
\Procedure{BuildSubspace}{$X$}
    \If{$|X| = 0$}
        \State \Return $(\varnothing, 0)$
    \EndIf
    
    \State Center $X$ using mean or median to get $X_c$
    \State Compute norms $\eta_t = \|X_{c,t}\|_2$
    \State Trim $\alpha$ fraction to get $X_k$
    \If{$|X_k| < 2$}
        \State \Return $(\varnothing, 0)$
    \EndIf
    
    \State Compute PCA with rank $r$ to obtain basis $U$
    
    \State Compute residuals $\rho(g^{(t)})$
    \State $\tau \gets \text{quantile}_q(\rho)$
    
    \State \Return $(U, \tau)$
\EndProcedure

\vspace{0.4em}
\State $(U, \tau) \gets \textsc{BuildSubspace}(X)$

\vspace{0.4em}
\For{$i=1$ to $n$}
    \If{$U \neq \varnothing$}
        \State $z_i \gets U^\top d_i$
        \State $\rho_i \gets \|d_i - U z_i\|_2$
    \Else
        \State $z_i \gets d_i$ \Comment{fallback}
        \State $\rho_i \gets 0$
    \EndIf
\EndFor

\vspace{0.3em}
\State $S \gets \text{KrumSelect}(\{z_i\}, f)$

\vspace{0.3em}
\State $G \gets \{ i\in S : \rho_i \le \tau \}$

\If{$|G| = 0$}
    \State $G \gets \{ \arg\min_i \rho_i \}$ \Comment{retain smallest residual}
\EndIf

\vspace{0.3em}
\State $a \gets \frac{1}{|G|}\sum_{i\in G} d_i$

\vspace{0.3em}
\State Append $a$ to buffer $X$ (evict oldest if needed)
\State \Return $a$

\end{algorithmic}
\end{algorithm}

\subsection{Hyperparameter Analysis}

Key hyperparameters:

\paragraph{$r$ (subspace dimension).}
Larger $r$ captures more benign variation but reduces spectral contrast with malicious updates. Range: 20 to 100 for CNNs.

\paragraph{$B$ (buffer size).}
Larger buffers improve stability but slow adaptation. $B=50$ balances these.

\paragraph{trim\_frac.}
Trimming 5 to 15\% of extremes stabilizes PCA.

\paragraph{$q$ (orthogonal energy quantile).}
Higher $q$ tightens the guard. $q = 0.98$ works well.

\paragraph{warmup\_rounds.}
Early rounds use median aggregation before enough history exists.

\paragraph{f\_byzantine.}
Krum's assumed adversary count.

\subsection{Why SpectralKrum Works}

SpectralKrum addresses existing weaknesses:

\begin{itemize}[leftmargin=*]
    \item \textbf{Non-IID drift:} projection aligns benign updates in spectral space.
    \item \textbf{Directional attacks:} spectral projection suppresses adversarial innovation directions.
    \item \textbf{Subspace-aware attacks:} orthogonal energy guard detects spectral mimicry failures.
    \item \textbf{Geometric stability:} Krum benefits from reduced dimensionality.
\end{itemize}

However, adaptive adversaries can mimic both spectral coordinates and orthogonal energy profiles, limiting effectiveness. This reveals fundamental limitations of spectral defenses.

\section{Attack Model and Adversarial Strategies}

Byzantine clients aim to corrupt the aggregated update. We consider a strong adversary: up to $f$ clients per round behave arbitrarily, constrained only to submit vectors of the correct dimensionality. Attacks may be adaptive (responding to the defense) or agnostic (fixed manipulations).

We study two attack families:

\begin{itemize}[leftmargin=*]
    \item \textbf{Classical model-poisoning}: simple heuristics like sign-flip, label-flip, min-max.
    \item \textbf{Spectrally informed}: attacks that exploit or mimic benign spectral structure, challenging PCA-based defenses.
\end{itemize}

We also evaluate a semantic backdoor attack.

We adopt a standard Byzantine threat setting in which up to $f$ out of $n$ clients per round are malicious. Malicious clients may collude and share information, possess white-box access to the global model, and attempt to reverse-engineer or estimate defense features such as PCA subspaces or orthogonal-energy thresholds; however, they do not have access to private client data from benign users. The adversary's goal varies by attack type: maximizing global loss (untargeted poisoning), forcing misclassification on certain classes (label-flip poisoning), implanting a targeted backdoor, steering the global model along controlled adversarial directions, or evading spectral and geometric outlier detection. SpectralKrum aims to provide robustness against this broad range of adversaries, including those that manipulate the spectral geometry of their updates. We now describe each attack in detail.

\subsection{Sign-Flip Attack}

The simplest attack negates the benign gradient~\cite{allen2020}: for benign update $b$, submit $d = -b$. This disrupts naive averaging and slows learning. Coordinate-wise defenses resist it, but Krum can be destabilized under non-IID dispersion.

Sign-flip is a baseline, not a sophisticated attack.

\subsection{Label-Flip Attack}

Malicious clients permute local labels~\cite{biggio2012} (e.g., $y \gets (y + 1) \bmod 10$) and train normally. The resulting gradients deviate from benign ones but may not produce large orthogonal energy, since they can remain aligned with benign principal components.

This attack tests spectral geometry limits: if malicious updates lie within the benign subspace, spectral filtering fails.

\subsection{Min-Max Attack}

The min-max attack~\cite{fang2020} maximizes deviation under norm constraints from benign statistics. With $\mu$ and $\sigma$ as the mean and variance of benign updates:

\[
d_{\text{minmax}} = \mu - c \sigma,
\]

for constant $c > 0$. The attack stays within benign variation range, often evading trimming and spectral guards.

This tests whether SpectralKrum can distinguish adversarial updates operating entirely inside the benign subspace.

\subsection{Buffer-Drift Attack (Subspace-Aware)}

This attack exploits SpectralKrum's historical buffer $X = \{g^{(1)}, \ldots, g^{(B)}\}$. The adversary gradually shifts the PCA subspace toward an adversarial direction by injecting small perturbations:

\[
d_{\text{mal}}^{(t)} = b^{(t)} + \epsilon \cdot v_{\text{adv}},
\]

where $b^{(t)}$ is the benign update, $v_{\text{adv}}$ is the adversarial direction, and $\epsilon$ increases slowly. By keeping $\epsilon$ below $\tau$, updates pass the guard while biasing the trajectory over time.

This challenges defenses relying on historical PCA structure.

\subsection{Adaptive-Steer Attack (Subspace-Aware)}

Adaptive-steer is the strongest attack in our suite and is specifically crafted 
to exploit spectral defenses. It follows a multi-step construction:

\paragraph{Step 1: Estimate the benign PCA subspace.}
The adversary computes or approximates the PCA basis $U_b$ from previously 
observed server broadcasts. Even partial or noisy estimates suffice.

\paragraph{Step 2: Proposed malicious direction.}
Let $v_{\text{adv}}$ denote the intended poisoning direction (e.g., negative 
benign mean direction or targeted class-steering direction).

\paragraph{Step 3: Projection into benign subspace.}
Compute:

\[
v_\parallel = U_b U_b^\top v_{\text{adv}}, \quad 
v_\perp = v_{\text{adv}} - v_\parallel.
\]

\paragraph{Step 4: Orthogonal-energy matching.}
Scale the orthogonal component to ensure it remains below the $\tau$ inferred 
from historical benign residuals:

\[
v_\perp' = \frac{\tau}{\|v_\perp\|_2 + \epsilon} \, v_\perp,
\]

where $\epsilon$ prevents division by zero.

\paragraph{Step 5: Final malicious update.}
Construct:

\[
d_{\text{adv}} = v_\parallel + v_\perp'.
\]

\paragraph{Why this evades spectral defenses.}

Because $v_\parallel$ lies inside the benign PCA subspace and $v_\perp'$ is 
carefully scaled to imitate benign orthogonal energy, the malicious update:

\begin{itemize}[leftmargin=*]
    \item appears spectrally consistent,
    \item matches benign orthogonal-energy statistics,
    \item evades PCA filtering in DnC-PMF, DnC-Cluster, and others.
\end{itemize}

This attack inspired SpectralKrum's guard mechanism, which applies Krum selection after projection and orthogonal energy filtering in the original space, combining geometric and spectral checks.

\subsection{Semantic Backdoor Attack}

Backdoor attacks modify behavior only on inputs with a trigger pattern. An adversary inserts trigger patches into local training images, relabels them to a target class, and trains gradients that associate trigger with target.

This is realistic for edge-device FL (e.g., stickers in phone images). Backdoors challenge aggregation defenses because gradients remain low-norm, orthogonal energy stays small, and the attack is task-specific rather than purely directional.

SpectralKrum provides partial resilience through geometric filtering but backdoors remain difficult.

\subsection{Why We Omit the Full DnC Attack}

The original DnC attack~\cite{shejwalkar2021} requires high coordination: malicious updates must be jointly optimized to form coherent clusters that confuse clustering-based detection. This needs synchronized adversaries, iterative optimization, and shared gradients, which is unrealistic for cross-device FL.

We instead evaluate a subspace-inspired stress-test capturing the key spectral properties: projection into benign subspace, orthogonal energy matching, and directional steering. This preserves the relevant geometric structure for evaluating SpectralKrum and DnC variants. Extending evaluation to the full DnC attack is an important direction for future research.

\subsection{Summary of Attack Suite}

We summarize all attacks and their properties in Table~\ref{tab:attacks}.

\begin{table}[H]
\centering
\caption{Summary of Attacks Evaluated}
\label{tab:attacks}
\begin{tabular}{p{2.3cm}|p{4.9cm}}
\toprule
\textbf{Attack} & \textbf{Description / Key Property} \\
\midrule
sign-flip & Negates benign gradient; high-norm, high variance \\
label-flip & Trains on corrupted labels; small orthogonal energy \\
min-max & Worst-case variance-based perturbation \\
buffer-drift & Steers PCA subspace over time \\
adaptive-steer & Subspace-aware, orthogonal-energy-matched \\
semantic backdoor & Trigger-based targeted misclassification \\
none (reference) & Baseline for comparison \\
\bottomrule
\end{tabular}
\end{table}

The diversity of this suite ensures that SpectralKrum is evaluated under 
classical, directional, spectral, and task-specific adversarial conditions.

\section{Experimental Setup}

This section details our evaluation environment, including datasets, model 
architectures, client sampling, non-IID partitioning, baselines, attacks, 
metrics, and implementation details. Our setup is designed to stress-test 
SpectralKrum under realistic and adversarial FL conditions.

\subsection{Datasets}

We use CIFAR-10~\cite{krizhevsky2009} (60,000 images, 10 classes), standard in FL research for its moderate difficulty and suitability for both shallow and deep models.

We partition data across $N=100$ clients using Dirichlet sampling~\cite{hsu2019} with $\alpha = 0.1$, creating highly skewed distributions where some clients see only a few classes. This non-IID setting amplifies geometric instability in benign gradients and stresses clustering-based defenses.

\subsection{Client Sampling}

Each round samples $n = 10$ clients uniformly without replacement from 100 total. Up to $f \leq 3$ may be Byzantine, representing typical cross-device FL.

\subsection{Model Architecture}

We use a lightweight CNN (TinyCNN) with two convolutional layers, max-pooling, and two fully connected layers. This architecture serves as a fast proxy for edge-device training while producing parameter vectors of moderate dimension ($d \approx 10^4$), sufficient to stress geometric and spectral defenses without the computational overhead of larger models.

\subsection{Baselines}

We compare against:

\begin{itemize}[leftmargin=*]
    \item \textbf{Trimmed Mean}: discards extreme coordinates.
    \item \textbf{Coordinate Median}: robust coordinate-wise estimator.
    \item \textbf{Geometric Median}: robust vector aggregation.
    \item \textbf{FullKrum}: classical single-update selection.
    \item \textbf{MultiKrum}: multi-update selection then averaging.
    \item \textbf{Bulyan}: iterated Krum with trimming.
    \item \textbf{DnC-PMF}: spectral divide-and-conquer with mixture filtering.
    \item \textbf{DnC-Cluster}: spectral clustering before aggregation.
\end{itemize}

These span coordinate-wise, geometric, and spectral defense philosophies.

\subsection{Hyperparameter Configuration}

SpectralKrum defaults:

\begin{itemize}[leftmargin=*]
    \item PCA dimension $r = 50$.
    \item Buffer size $B = 50$.
    \item Centering: mean.
    \item Trim mode: two-sided.
    \item Trim fraction: $0.1$.
    \item Warmup rounds: $3$.
    \item PCA refresh interval: $1$.
    \item Orthogonal-energy quantile: $q = 0.98$.
    \item Minimum guard survivors: $1$.
    \item Krum parameter $f$: assumed number of Byzantine clients.
\end{itemize}

These values are derived from empirical stability tests and provide a robust 
balance between PCA estimation quality and computational efficiency.

\subsection{Training Procedure}

Clients perform $E=1$ or $E=2$ local epochs with mini-batch SGD~\cite{bottou2010} (learning rate 0.01, weight decay $5 \times 10^{-4}$) and data augmentation (random crop, horizontal flip) for benign clients. Malicious clients modify training or updates per attack.

\subsection{Evaluation Metrics}

\textbf{Per-Round Accuracy}: global model accuracy during training.

\textbf{AUC}: $\frac{1}{T} \sum_{t=1}^T \text{Acc}(t)$, measuring stability across rounds.

\textbf{Best/Final Accuracy}: peak and convergence accuracy.

\textbf{Computation Overhead}: PCA, projection, Krum, and guard filtering times.

\subsection{Implementation Details}

SpectralKrum is implemented in Python with NumPy and scikit-learn~\cite{pedregosa2011}. Experiments run on an NVIDIA RTX 4090 GPU (training) and CPU (aggregation). PCA overhead is negligible relative to forward/backward passes. We use fixed random seeds for reproducibility.

\subsection{Summary}

Our setup evaluates SpectralKrum under highly non-IID data, diverse attacks, strong adversaries, and eight robust baselines.

\section{Results}

We evaluate SpectralKrum across seven adversarial settings and nine aggregation methods (eight baselines plus SpectralKrum). Our focus is on understanding robustness under challenging scenarios, especially when benign gradients are non-IID and adversarial updates intentionally exploit spectral structure.

We present a compact \emph{AUC} summary (mean accuracy across rounds) to rank 
defenses, followed by per-attack accuracy curves that illustrate the temporal 
behavior underlying those summary scores. For backdoor attacks we also discuss
attack success dynamics qualitatively.

\subsection{AUC summary (mean accuracy across rounds)}

Table~\ref{tab:auc_summary} reports the mean accuracy across rounds (an
AUC-like summary) at attacker\_count = 2 for each (algorithm, attack) pair. These
values are computed by averaging accuracy across training rounds and then
averaging across experimental repeats and seeds. Higher values indicate better
average robustness during training.

\begin{table*}[t]
\centering
\caption{Mean accuracy across rounds (AUC-like) at attacker\_count = 2.
(Each cell is the mean accuracy \textit{across rounds} for the given
(algorithm, attack) pair.)}
\label{tab:auc_summary}
\small
\begin{tabular}{lrrrrrrrr}
\toprule
Algorithm & adaptive\_steer & buffer\_drift & label\_flip & min\_max & none & semantic\_backdoor & sign\_flip & Grand Total \\
\midrule
Bulyan & 40.7727 & 33.7527 & 37.2818 & 41.4727 & 38.9691 & 42.5200 & 41.3651 & 39.4596 \\
CoordMedian & 35.7364 & 44.8918 & 47.7245 & 45.4373 & 47.9509 & 48.4664 & 43.5873 & 44.8273 \\
DnC-Cluster & 24.7427 & 47.1036 & 50.3664 & 47.1200 & 46.5409 & 52.4809 & 19.7309 & 39.7654 \\
DnC-PMF & 51.1427 & 49.8391 & 50.4727 & 50.8836 & 51.6400 & 51.1864 & 50.9018 & 50.8673 \\
FullKrum & 26.3655 & 27.3945 & 27.2727 & 26.4182 & 26.7818 & 27.8309 & 26.3891 & 26.3793 \\
GeometricMedian & 39.5818 & 54.6645 & 57.1027 & 56.4127 & 57.8000 & 57.5309 & 52.0718 & 53.1380 \\
MultiKrum & 50.9555 & 47.4682 & 50.4355 & 50.8391 & 51.6655 & 50.6691 & 50.8864 & 50.4170 \\
SpectralKrum & 49.7164 & 47.5145 & 38.9818 & 48.7055 & 44.4955 & 47.1764 & 49.7318 & 46.4317 \\
TrimmedMean & 45.2473 & 53.1673 & 55.7927 & 51.7682 & 55.1891 & 55.4691 & 49.0585 & 52.3549 \\
\midrule
\textbf{Grand Total (mean over algos)} & \textbf{41.6739} & \textbf{45.0885} & \textbf{46.2052} & \textbf{48.5798} & \textbf{46.7258} & \textbf{48.1478} & \textbf{41.6505} & \textbf{47.6238} \\
\bottomrule
\end{tabular}
\end{table*}

The table provides a concise ranking: methods such as GeometricMedian,
MultiKrum, DnC-PMF, and TrimmedMean achieve high average accuracy across 
many attacks, while FullKrum consistently yields low accuracy for these 
non-IID settings. However, averages obscure temporal dynamics: the following 
subsections show round-by-round accuracy curves that reveal stability, 
oscillation, and convergence behavior that explain the AUC scores.

\subsection{Adaptive-Steer (accuracy curve)}

\begin{figure}[H]
\centering
\includegraphics[width=\linewidth]{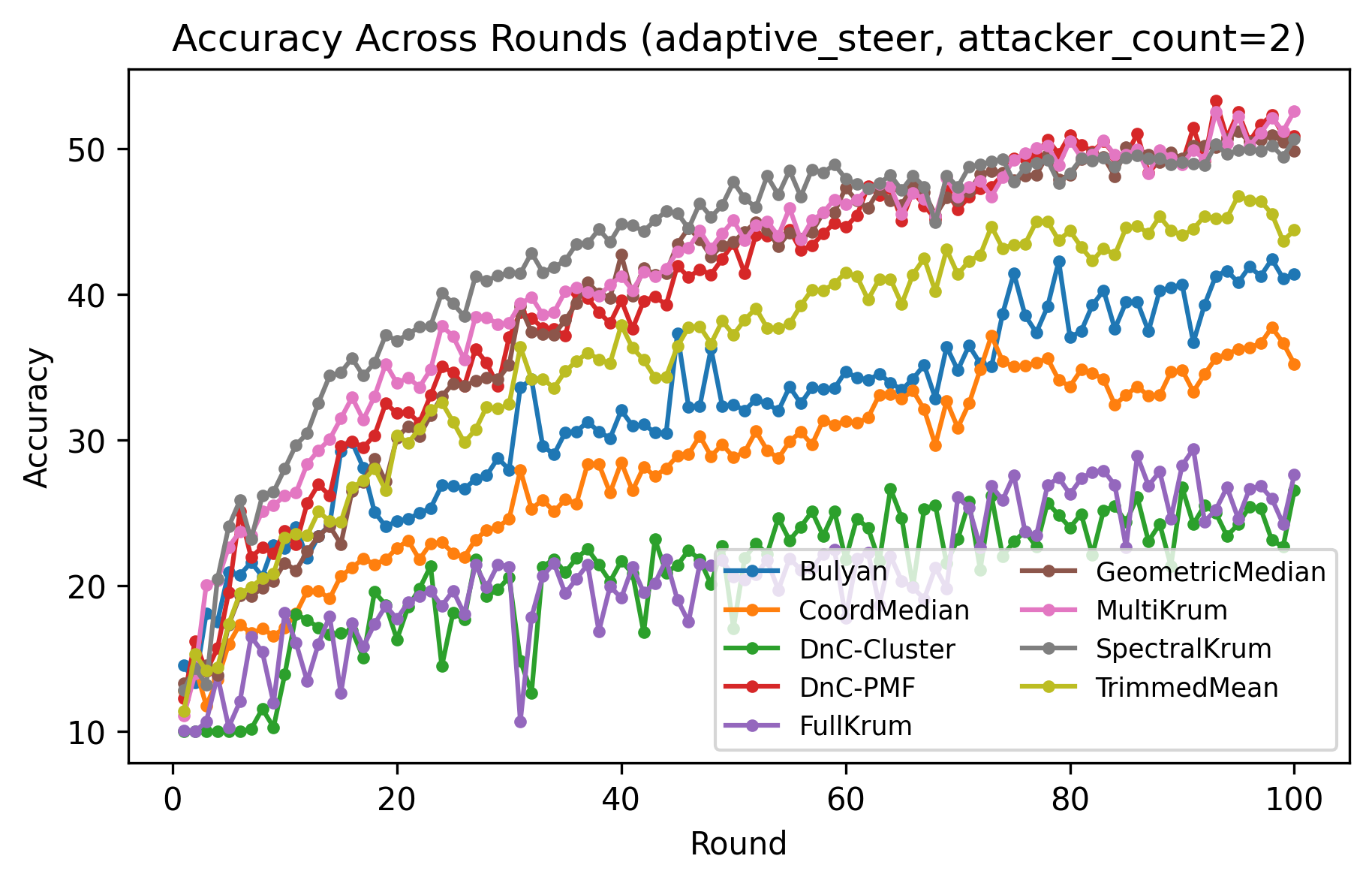}
\caption{Accuracy across rounds under adaptive\_steer (attacker\_count = 2).}
\label{fig:acc_adaptive}
\end{figure}

Adaptive-steer is a subspace-aware attack that tries to remain spectrally
consistent with benign updates while steering the global model. The
accuracy curve in Figure~\ref{fig:acc_adaptive} shows how different
defenses evolve over time under this attack (attacker\_count = 2).

From the curve we observe that SpectralKrum and DnC-PMF / MultiKrum 
achieve higher accuracy quickly and remain relatively stable, whereas 
methods such as FullKrum and DnC-Cluster show pronounced oscillation 
or persistently lower accuracy. This temporal pattern aligns with the 
AUC entries in Table~\ref{tab:auc_summary}.

\subsection{Buffer-Drift (accuracy curve)}

\begin{figure}[H]
\centering
\includegraphics[width=\linewidth]{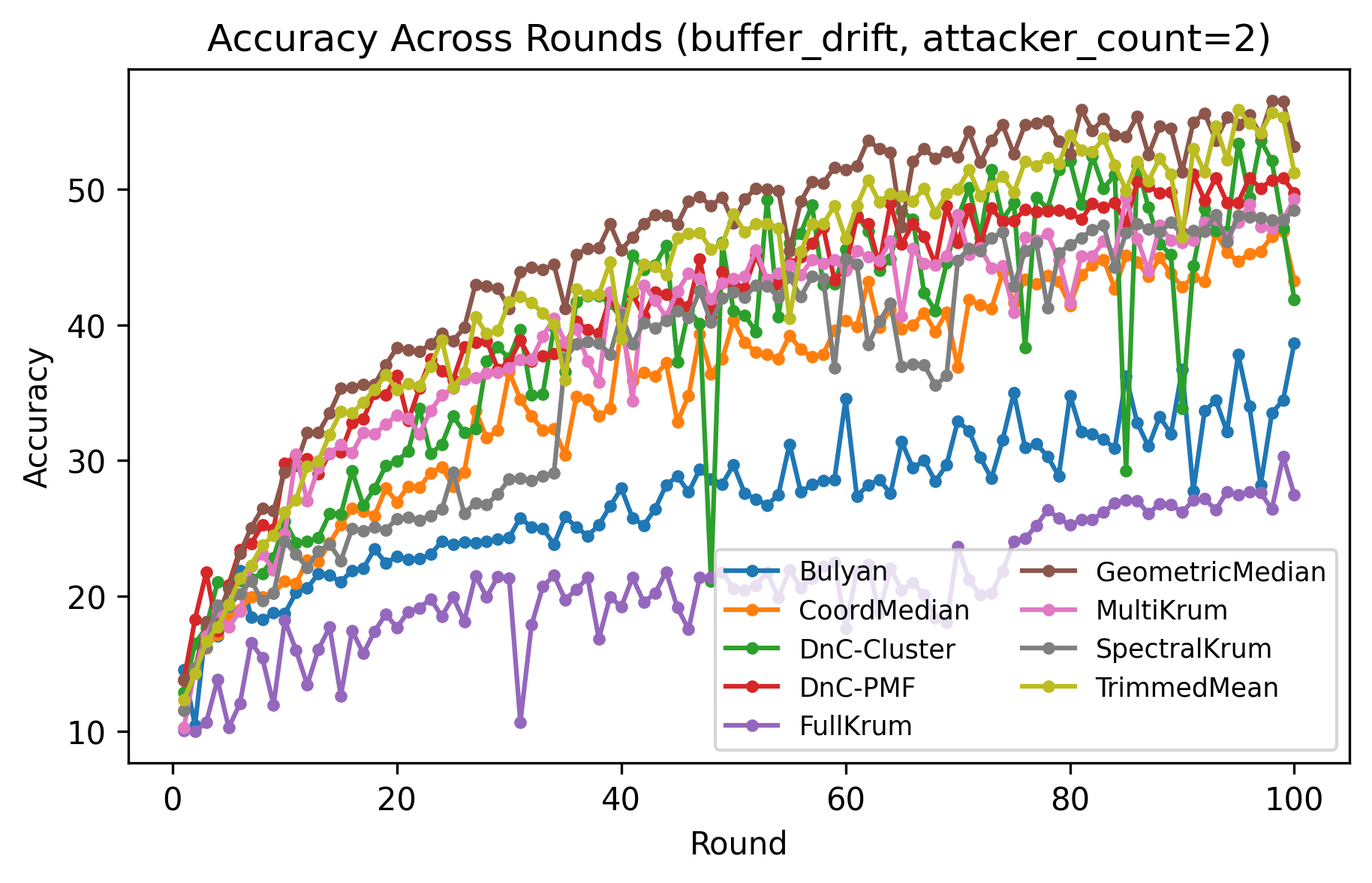}
\caption{Accuracy across rounds under buffer\_drift (attacker\_count = 2).}
\label{fig:acc_buffer_drift}
\end{figure}

Buffer-drift targets the historical PCA buffer and attempts gradual subspace
manipulation. The curves show that GeometricMedian, TrimmedMean, and several
median-like aggregators remain robust and converge to high accuracy. Krum-based
variants that do not leverage robust spectral filtering (e.g., FullKrum) lag
behind. SpectralKrum tracks the top performers closely, illustrating its
temporal resilience to drift.

\subsection{Label-Flip (accuracy curve)}

\begin{figure}[H]
\centering
\includegraphics[width=\linewidth]{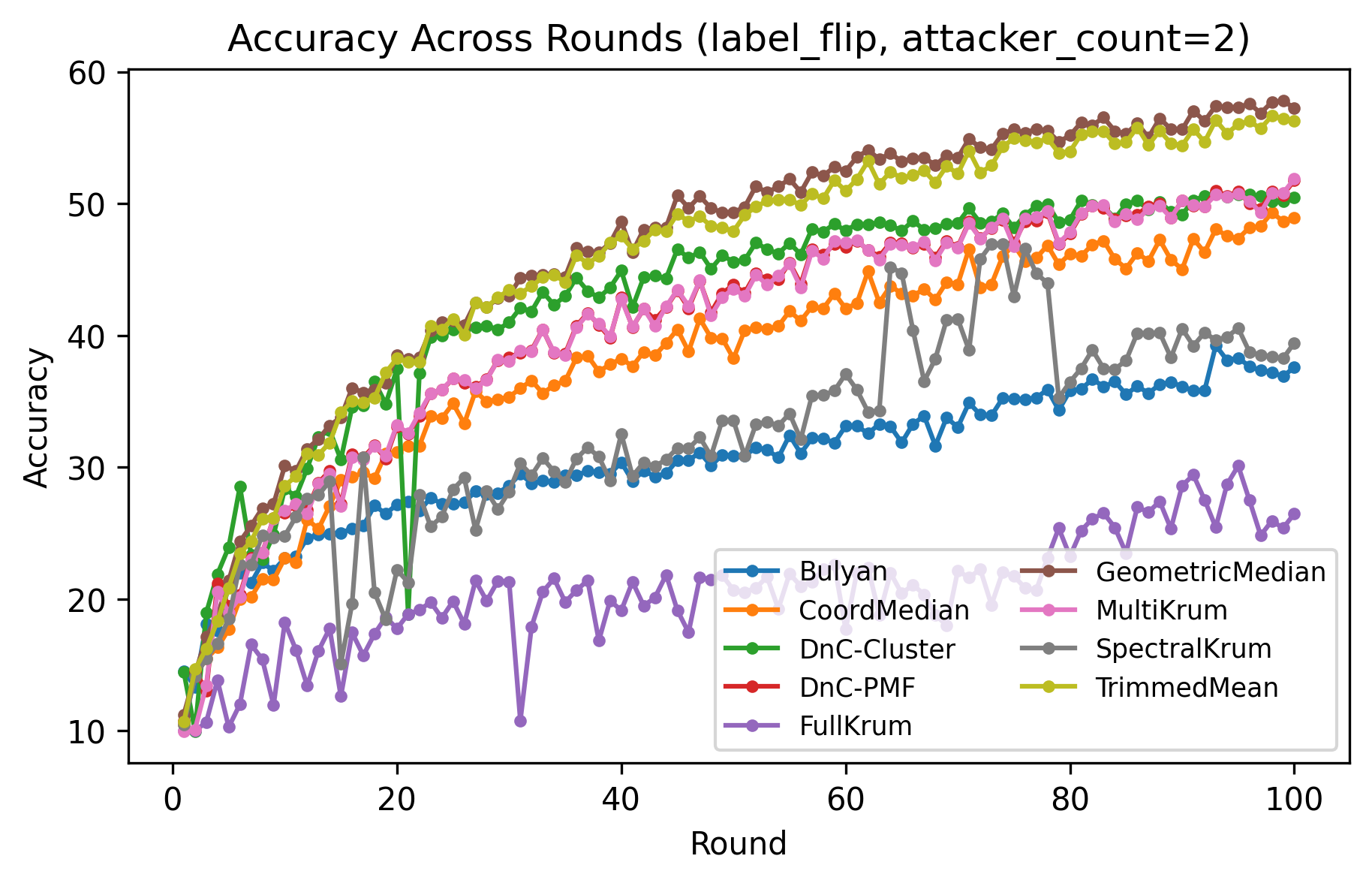}
\caption{Accuracy across rounds under label\_flip (attacker\_count = 2).}
\label{fig:acc_label_flip}
\end{figure}

Label-flip poisoning changes local labels and therefore injects semantically
meaningful but spectrally subtle gradients. Many coordinate-robust or
median-based methods (TrimmedMean, GeometricMedian) obtain strong final
accuracy because their coordinate-level robustness mitigates systematic 
label noise. SpectralKrum performs well on average but is less dominant 
here because label-flip updates can remain largely inside the benign 
subspace, reducing the orthogonal-energy signal.

\subsection{Min-Max (accuracy curve)}

\begin{figure}[H]
\centering
\includegraphics[width=\linewidth]{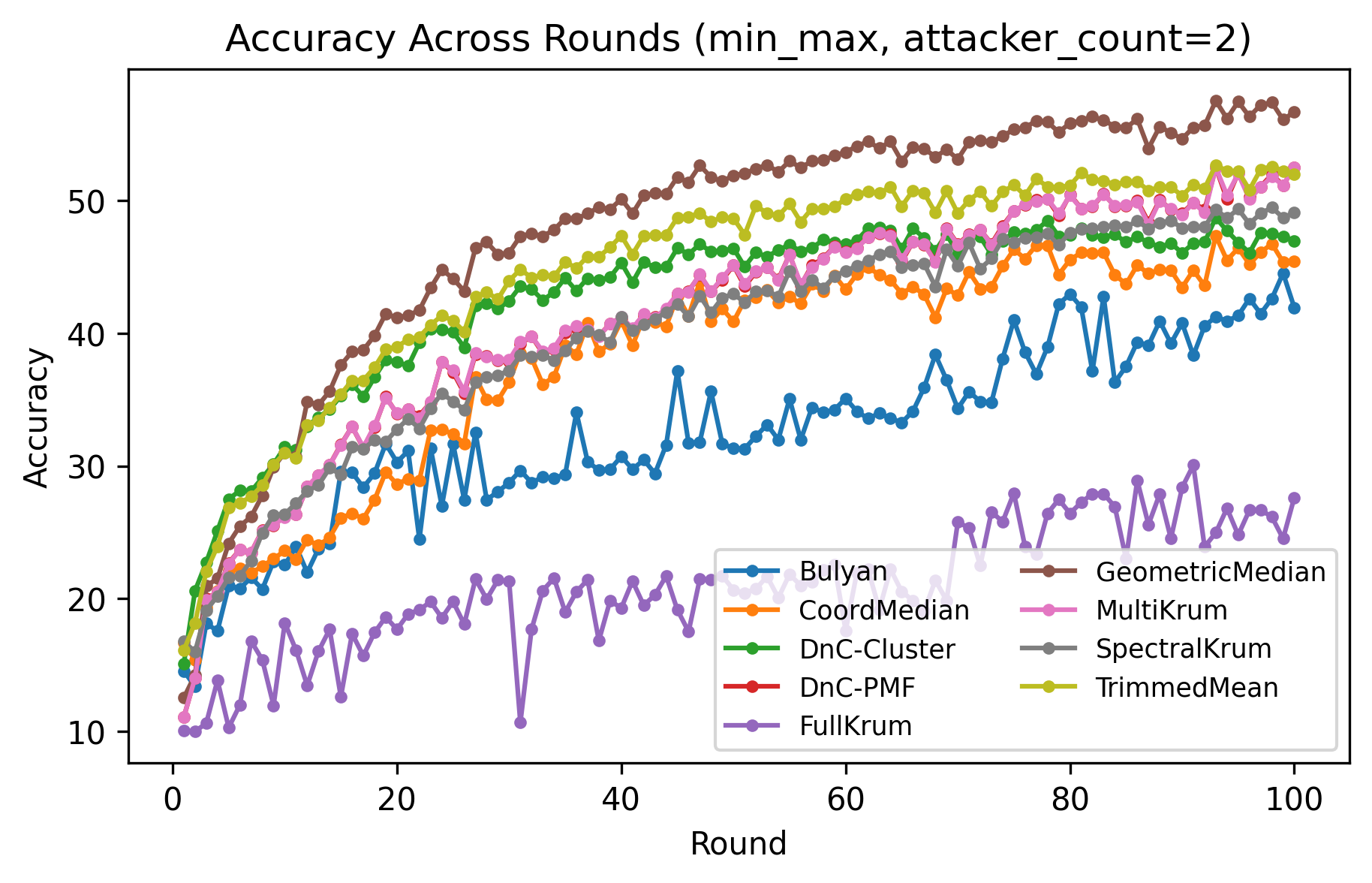}
\caption{Accuracy across rounds under min\_max (attacker\_count = 2).}
\label{fig:acc_min_max}
\end{figure}

The min-max attacker attempts worst-case perturbations within benign variation.
Accuracy curves show that spectral defenses offer modest advantage, but the
attack remains hard to detect solely via orthogonal energy. DnC-PMF, MultiKrum,
and trimmed/median methods remain highly competitive.

\subsection{Semantic Backdoor (accuracy curve)}

\begin{figure}[H]
\centering
\includegraphics[width=\linewidth]{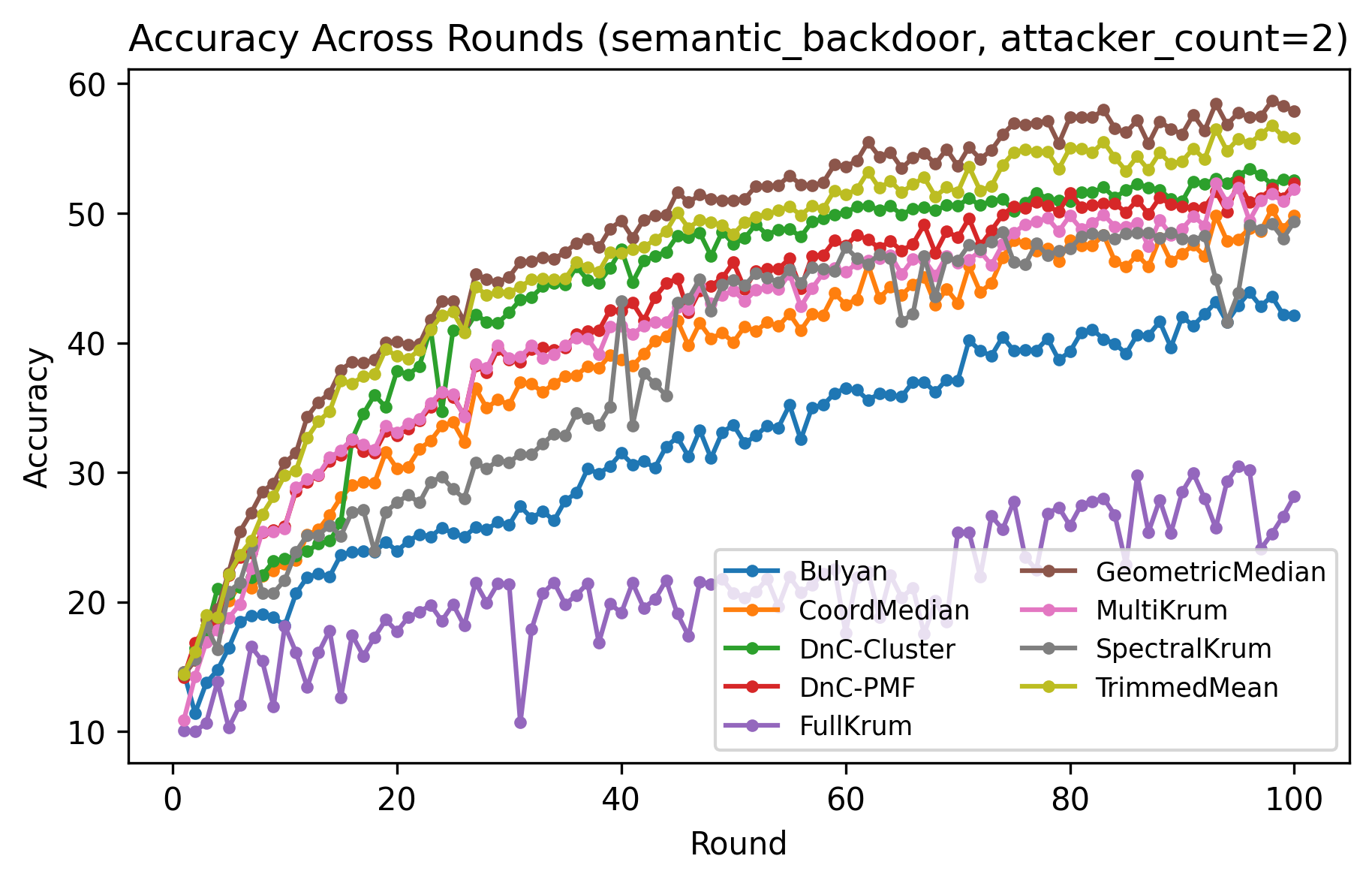}
\caption{Accuracy across rounds under semantic\_backdoor (attacker\_count = 2).}
\label{fig:acc_backdoor}
\end{figure}

Backdoor attacks produce gradients targeted toward a trigger-based task 
manipulation and often exhibit low orthogonal energy. Consequently, most 
aggregation-only defenses, including SpectralKrum, reduce but do not fully 
eliminate backdoor ASR. The AUC table highlights which defenses maintain 
high overall accuracy in the presence of backdoor clients, while 
Figure~\ref{fig:acc_backdoor} shows that some defenses converge more smoothly 
than others.

\subsection{Sign-Flip (accuracy curve)}

\begin{figure}[H]
\centering
\includegraphics[width=\linewidth]{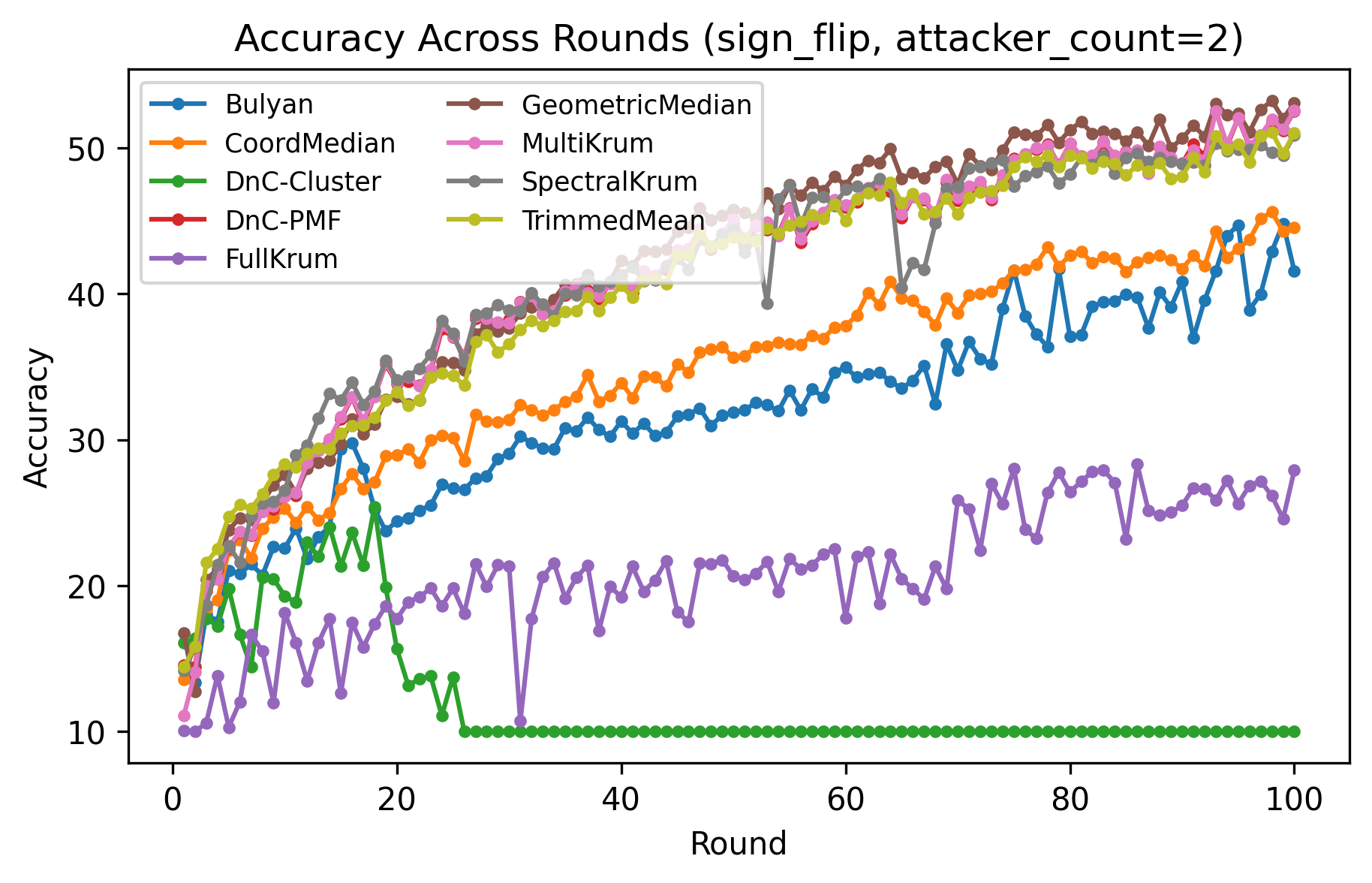}
\caption{Accuracy across rounds under sign\_flip (attacker\_count = 2).}
\label{fig:acc_sign_flip}
\end{figure}

Sign-flip is a high-magnitude but geometrically simple untargeted attack, and 
most robust aggregation rules detect it effectively. Methods such as 
GeometricMedian, MultiKrum, TrimmedMean, DnC-PMF, and SpectralKrum rise 
steadily to high accuracy, whereas cluster-based defenses such as DnC-Cluster 
collapse early and remain stuck at chance level. Spectral projection helps 
stabilize variance, and in combination with median-style robustness keeps 
the global model accurate despite the attack.

\subsection{Computational Overhead}

Table~\ref{tab:overhead_numeric} reports the measured mean total computation
time per training round for each aggregation rule. These values were collected
under identical hardware and software conditions and therefore provide a fair
comparison of relative cost across methods.

\begin{table}[H]
\centering
\caption{Mean total computation time per round for each aggregation rule.}
\label{tab:overhead_numeric}
\begin{tabular}{l|r}
\toprule
Aggregation Rule & Mean Time (ms) \\
\midrule
FullKrum       & 7.111224 \\
MultiKrum      & 8.240192 \\
CoordMedian    & 29.384039 \\
GeometricMedian & 76.383777 \\
DnC-PMF        & 167.312114 \\
DnC-Cluster    & 368.234553 \\
Bulyan         & 401.891775 \\
TrimmedMean    & 406.241334 \\
SpectralKrum   & 1311.667696 \\
\bottomrule
\end{tabular}
\end{table}

Krum-based methods introduce modest overhead from distance computations.  
Median-based approaches (CoordMedian, GeometricMedian) are significantly 
more expensive.  
DnC-style methods incur very high costs due to clustering or matrix 
factorization.  
\textbf{SpectralKrum is the most expensive} because each round performs 
PCA-based subspace construction, projection, and orthogonal-energy filtering, 
but the cost remains practical in server-side federated learning deployments.

\section{Discussion}

The experimental results paint a nuanced picture that resists simple conclusions. SpectralKrum neither dominates existing methods nor fails completely; its effectiveness depends critically on the attack's geometric signature.

\subsection{When Spectral Geometry Helps}

SpectralKrum performs competitively against attacks that introduce detectable spectral anomalies. Under adaptive-steer and buffer-drift, SpectralKrum matches or exceeds DnC-PMF and MultiKrum. Even subspace-aware adversaries often introduce residual orthogonal energy from imperfect subspace estimation or inherent attack constraints.

\subsection{When Spectral Geometry Fails}

Label-flip and min-max attacks produce gradients that remain within the benign subspace. Their malicious effect comes from direction, not orthogonal deviation. Against such attacks, SpectralKrum's guard provides no signal. Table~\ref{tab:auc_summary} shows GeometricMedian and TrimmedMean outperforming SpectralKrum on label-flip by substantial margins. Coordinate-wise methods succeed because they aggregate each parameter independently, diluting systematic perturbations.

\subsection{The Semantic Gap}

Semantic backdoors present an even deeper challenge: their gradients are geometrically indistinguishable from benign training on a modified distribution. No aggregation-only defense can fully neutralize well-crafted backdoors. This suggests robust FL requires layered defenses: gradient-level filtering for geometric anomalies combined with activation-level analysis for semantic manipulations.

\subsection{Limitations}

Several directions emerge:

\paragraph{Robust or Incremental PCA.}
Under non-IID conditions, standard PCA becomes unstable. Robust PCA or streaming variants could maintain cleaner subspace estimates.

\paragraph{Adaptive Rank Selection.}
Automatically choosing PCA dimension $r$ based on eigengaps or variance retention may prevent underfitting or overfitting the spectral subspace.

\paragraph{Hybrid Spectral-Median Aggregation.}
Coordinate-wise statistics (Trimmed Mean, Geometric Median) outperform spectral filters on label-flip and min-max. A hybrid combining spectral projection with coordinate-wise median may inherit complementary strengths.

\paragraph{Layer-wise SpectralKrum.}
Deep networks have heterogeneous layers with different geometric structures. Layer-wise or multi-subspace PCA may capture richer variation.

Our evaluation has inherent limitations. We tested only CIFAR-10 with a TinyCNN; real deployments involve diverse modalities where gradient spectral structure may differ substantially. We evaluated fixed adversary fractions ($f \leq 3$ of $n=10$); higher fractions and larger pools could reveal different dynamics. Our adversaries are static; fully adaptive adversaries tracking the server's PCA in real-time might defeat SpectralKrum more effectively.

\section{Conclusion}

SpectralKrum combines spectral subspace estimation with Krum's geometric selection, aiming to exploit the low-dimensional structure of benign FL trajectories while filtering spectrally anomalous updates. Our extensive evaluation across seven attack families and eight robust baselines reveals a clear pattern: spectral defenses help when attacks introduce orthogonal variance but provide no advantage when malicious gradients lie within the benign subspace.

This finding has broader implications for Byzantine-robust FL. No single defense philosophy (coordinate-wise, geometric, or spectral) addresses all attack families. The path forward likely involves hybrid architectures that combine complementary signals, perhaps applying spectral filtering to detect geometric anomalies while using activation-level analysis for semantic attacks that evade gradient-based detection.

We release our code and experimental logs to facilitate reproducibility and encourage further investigation into the interplay between spectral structure and Byzantine robustness in federated optimization.

\end{document}